\documentclass[conference]{IEEEtran}
\IEEEoverridecommandlockouts
\usepackage{cite}
\usepackage{amsmath,amssymb,amsfonts}
\usepackage{algorithmic}
\usepackage{graphicx}
\usepackage{textcomp}
\usepackage{xcolor, multirow, adjustbox}
\def\BibTeX{{\rm B\kern-.05em{\sc i\kern-.025em b}\kern-.08em
    T\kern-.1667em\lower.7ex\hbox{E}\kern-.125emX}}
\begin{document}

\title{Universal Recurrent Event Memories for Streaming Data

\thanks{This work was partially supported by ARO grant W911NF-21-1-0254.}
}

\author{\IEEEauthorblockN{1\textsuperscript{st} Ran Dou}\\
\IEEEauthorblockA{\textit{Department of Electrical and Computer Engineering} \\
\textit{University of Florida}\\
Gainesville, FL, United States \\
dour@ufl.edu}
\and
\IEEEauthorblockN{2\textsuperscript{nd} Jose Principe}
\IEEEauthorblockA{\textit{Department of Electrical and Computer Engineering} \\
\textit{University of Florida}\\
Gainesville, FL, United States \\
principe@cnel.ufl.edu}
}
\maketitle

\begin{abstract}
In this paper, we propose a new event memory architecture (MemNet) for recurrent neural networks, which is universal for different types of time series data such as scalar, multivariate or symbolic. Unlike other external neural memory architectures, it stores key-value pairs, which separate the information for addressing and for content to improve the representation, as in the digital archetype. Moreover, the key-value pairs also avoid the compromise between memory depth and resolution that applies to memories constructed by the model state. One of the MemNet key characteristics is that it requires only linear adaptive mapping functions while implementing a nonlinear operation on the input data. MemNet architecture can be applied without modifications to scalar time series, logic operators on strings, and also to natural language processing, providing state-of-the-art results in all application domains such as the chaotic time series, the symbolic operation tasks, and the question-answering tasks (bAbI). Finally, controlled by five linear layers, MemNet requires a much smaller number of training parameters than other external memory networks as well as the transformer network. The space complexity of MemNet equals a single self-attention layer. It greatly improves the efficiency of the attention mechanism and opens the door for IoT applications.
\end{abstract}


\section{Introduction}
Time series analysis has been widely studied for years and continues to be an active research area. State models have been the norm to extract information from time series.  From autoregressive moving average (ARMA) \cite{akaike1973maximum, benjamin2003generalized}, hidden Markov model(HMM)  \cite{hmm, schuller2003hidden} to recurrent neural networks (RNNs) \cite{doi:10.1073/pnas.79.8.2554, giles1992learning}, how to preserve and propagate hidden information is an essential problem. The simplest and easiest to train memory model is the moving average (MA) while the autoregressive (AR) models use recurrency and are more efficient (fewer parameters). The gamma memory \cite{de1992gamma} studied the fundamental  properties of linear memory (the compromise between memory depth and resolution), while RNNs employ nonlinear memories, with a substantial increase in computation complexity and bringing new problems such as vanishing gradients. HMMs assume a set of unobservable states in the dynamic system to model transitions. However, all these models have the same problem, the lack of representing long-term dependencies. Taking symbolic time series as an example, the prediction is not only affected by the current input sample and previous hidden state but is also affected by the knowledge of its history. Therefore, gated recurrent neural networks have been proposed \cite{hochreiter1997long, cho2014properties}. By controlling the switches for an external hidden state, they create a representation linking the current to past samples with arbitrary lags. These methods greatly improve the processing of long-term dependencies for time series models and they can be considered the first neural network models with external memory. However, there are still problems with gated networks. When propagating the hidden information, the states at different times are merged into one representative state. When the external representation is updated, past information is forgotten. One solution to this problem is to store the external information separately in multiple memory locations, leading to many external memory architectures \cite{ma2019taxonomy} such as Neural Stack \cite{grefenstette2015learning}, NTM \cite{graves2014neural}, DNC \cite{graves2016hybrid},
DMN \cite{kumar2016ask} etc. These memory architectures are adapted and the read and write operations must be constructed using continuous functions, which create resolution and redundancy issues that are hard to optimize \cite{sun2017neural}.  MemN2N \cite{sukhbaatar2015end} and MEMO \cite{banino2020memo} combined the external memory with the attention mechanisms, which does not employ a recurrent system.

The concepts of query, key and value were first proposed in the transformer network \cite{vaswani2017attention} and the attention mechanism has been a hot topic in natural language processing. Key and value have a direct relation with digital computer memories implemented with an address (key) and the memory content (value), but they are utilized in a learning framework. The long-term information can be well retrieved by querying the key and value matrices, however, transformer networks do not use any state representation, which is known to be very efficient for storing past information. Inspired by this architecture, we propose in this paper a novel architecture for recurrent networks with an external neural memory that can be accessed in a similar manner and constitutes a new element in the external memory taxonomy presented in \cite{ma2019taxonomy}. This new event memory architecture will be called memory net (MemNet), where the event information at each time is separated into two vectors, one for the addressing and another for the content. Controlled by a single-layer linear recurrent model, MemNet can achieve state-of-the-art performance with a much smaller number of trainable parameters when compared with models using nonlinearities on different tasks such as chaotic time series prediction, logic operator tasks and question answering dataset (bAbI) in natural language processing. The paper shows that the key-value decomposition of the memory event is free from the memory depth resolution trade-off and also avoids the precision efficiency bottleneck created by the implementation of continuous read and write operators.  

\section{A Brief Review of External Neural Memory}

The use of external neural memory is becoming more popular. Taking NTM as an example, an external memory $M$ takes the form of,
\begin{equation}
    M = \left[
    \begin{array}{cc}
         M(1)  \\
         M(2)  \\
         ...  \\
         M(n)
    \end{array}
    \right] \in R^{n\times m}
\end{equation}
where $M(i)$ is the $i$th item stored in the memory with a size of $m$ bits, and $n$ is the memory size, which limits the number of items that can be stored in the memory. There are two main operations in the external neural memory: read and write. The read operation copies the memory contents and the write operation includes adding new items and erasing existing items. The controllers that emit commands to the memory are called read and write heads. Unlike the conventional read-and-write operations in digital computers, external memory in machine learning operate with continuous values and the operators themselves are smooth functions. We present here their most common implementation \cite{graves2014neural}.
\subsection{Read operation}
For the read operation, the read head emits a set of weights $w^r_i$ for each location, and for normalization, the sum of these weights equals to 1.
\begin{equation}
    \sum_{i=1}^n w^r_i = 1
\end{equation}

The read vector $r_t$ is defined by the weighted combination of memory items $M_t(i)$ at location $i$.
\begin{equation}
    r_t = \sum_{i=1}^n w^r_i M_t(i)
\end{equation}
Notice that this is not a very effective implementation because the value of the memory $M_t(i)$ is multiplied by a vector of constant norm, which is a convolution operation. hence, small variations in $M_t(i)$ can be lost in the read, affecting the precision of the operator.  
\subsection{Write operation}
The first operation for write is an erase. At the time $t$, the write head emits an erase weights $w^e_t$. The erased memory is updated by,
\begin{equation}
    \widetilde M_t (i) = M_t(i) (1 - w_t^e(i))
\end{equation}
where $\widetilde M_t (i)$ is the memory after erasing.
Then, the write head emits a value vector $v_t$, which needs to be stored in the memory and a weight vector $w_t^w$ for the locations.
\begin{equation}
    M_{t+1}(i) = \widetilde M_t (i) + w_t^w(i)v_t
\end{equation}

\subsection{Addressing mechanisms}
Performing the read and write operations requires an addressing mechanism. There are two types of addressing mechanisms for neural memories that need to be used concurrently \cite{graves2014neural}: content-based addressing and location-based addressing. For content-based addressing, a query vector $q_t$ is emitted by the heads at time $t$. Then, the weight $w^c_t(i)$ is defined by the similarity between $q_t$ and $M_t(i)$. Using $K(\cdot, \cdot)$ to denote the similarity measure,
\begin{equation}
    w^c_t(i) = K(M_t(i), q_t)
\end{equation}
For location-based addressing, a scalar interpolation gate $g_t\in(0, 1)$ is emitted by the heads. Then, the weight is updated by,
\begin{equation}
\label{weight}
    w_t = g_t w_t^c + (1 - g_t)w_{t-1}
\end{equation}

\subsection{Shortcomings}
The external memory was proposed to mimic computer memory operations. However, there are shortcomings in this implementation. Conceptually, the external memory in the computer is a set of physical addresses. It uses pointers (memory addresses) to retrieve corresponding values. In neural networks, to make this operation differentiable, the memory item is used for both addressing and content which are totally different types of information in computer memories. This limits the precision of the memory as well as its capacity.

The second shortcoming is theoretical. State memory architectures suffer from the trade-off between the memory depth and the resolution. For a recurrent system, the information decays at a certain rate $\mu \in (0, 1)$ when propagating through time. A small decay rate means that the information decays slowly and has been smoothed in time, which causes a poor time resolution because local samples are averaged together. The memory depth $d$ is the furthest sample that the system can gain information. It has been shown that the product of the resolution and the memory depth for linear memories is a constant given by the filter order \cite{193206}. The tradeoff also exists for nonlinear state models as the vanilla RNN but the product of memory and resolution is much harder to determine analytically. For external memory, the memory depth can be improved, but at the sacrifice of resolution. Therefore, when the read head emits a query vector $q_t$, it can only retrieve items that are similar to it. The system gains little from these because the read content $r_t$ and the query vector $q_t$ contain similar information. Therefore, external memories still suffer from the trade-off between the resolution and the memory depth.

In order to solve these two fundamental problems, our proposal is to separate the memory event into key and value and design two independent sub-modules: one is the addressing system, where a low decay rate system provides a good memory depth that helps with the long-term dependency. It behaves like the physical addresses on the computer. According to the trade-off mentioned above, it has a bad resolution in time, where the information is been smoothed but we will not use it for value retrieval. Another sub-module is the read content, where a fast decay system is used to provide a good temporal resolution for information. This sub-module behaves like the content stored in the computer memory that lasts forever. Naturally, we can operate with the two sub-modules in pairs and retrieve information as computer programs do.
\\
\section{Advantage of Key-value Pairs and Nonlinear Mapping}

Given a set of input features $\{x_1, x_2, ..., x_n\}$ and the corresponding desired outputs $\{y_1, y_2, ..., y_n\}$, the goal of a feed-forward neural network is to find the weights $\theta$ for the nonlinear mapping function $f_\theta(\cdot)$ that provides the minimum prediction error.
\begin{equation}
    \theta_{opt} = \min_\theta E(loss(f_\theta(x_i), y_i))
\end{equation}
For the universality of the mapping functions, many methods have been proposed such as multilayer perceptrons (MLPs). However, these neural networks are too complex and work as a black box. Because the proposed architecture separates addresses from contents, only the content sub-module needs to be nonlinear. Moreover, because of time resolution, we can use a local nonlinearity such as the Gaussian kernel to simplify the nonlinear mapping. 
\begin{figure}[h]
    \centering
    \includegraphics[scale=0.5]{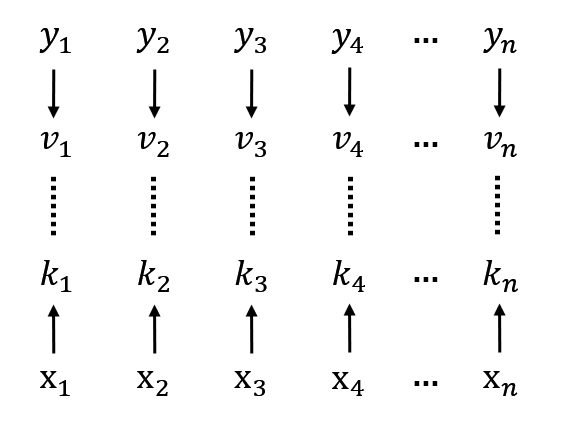}
    \caption{Key-value pairs and nonlinear mapping.}
    \label{fig:key-value pairs}
\end{figure}

As shown in Figure \ref{fig:key-value pairs} for a classification task, we use the keys $\{k_1, k_2, ..., k_n\}$ to capture the input features and the values $\{v_1, v_2, ..., v_n\}$ to store the corresponding values from the desired output. When a new input sample arrives, the Gaussian similarity is used to select the pairs based on the keys and provide the output based on the values. For example, given a set of features $\{1, 2, 3, 4\}$ for the input and $\{1, -1, 1, -1\}$ for the corresponding labels an MLP needs at least one hidden layer with nonlinear activation function for classification. However, this can be easily done by creating the key-value pairs $\{(k, v): (1, 1), (2, -1), (3, 1), (4, -1)\}$. And the nonlinear mapping can be expressed as,
\begin{equation}
    f(x) = G_\sigma(x, 1) - G_\sigma(x, 2) + G_\sigma(x, 3) - G_\sigma(x, 4)
\label{eq:eval}
\end{equation}
where $\sigma$ can be set as a small value to make the Gaussian close to an impulse. Compared with the black-box hidden layer operations, the key-value pair architecture implements the mapping in a much simpler and concise manner and is easy to understand as well. Also, since the similarity is only evaluated by the data points, the key-value pair architecture tends to be zero for out-of-distribution data. In this way, the nonlinear mapping from inputs to outputs can be easily built by a key-value pair architecture.

\begin{figure*}[t]
    \centering
    \includegraphics[scale=0.58]{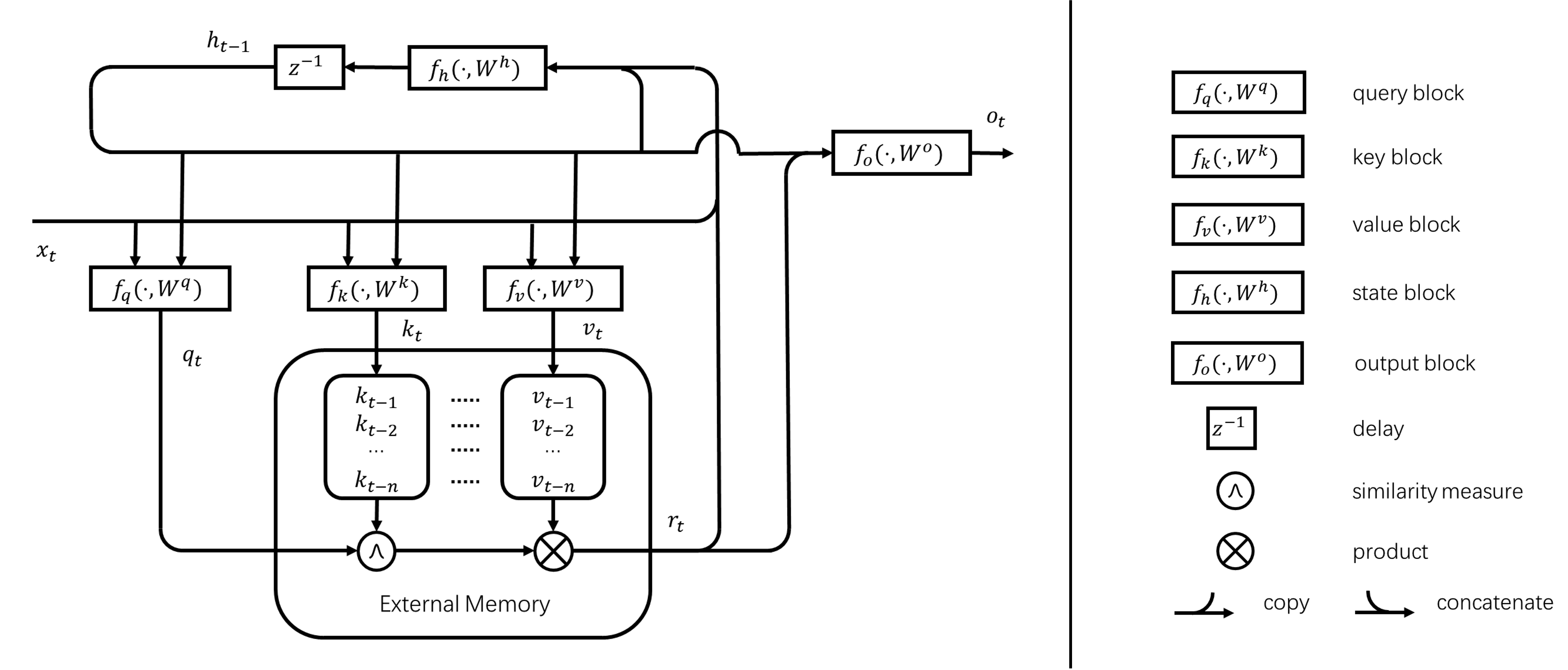}
    \caption{MemNet Architecture. In the external memory, each key vector $k_i$ is compared the similarity with the query vector $q_t$, then the read vector $r_t$ is the weighted sum of the values in the memory.}
    \label{fig:Architecture}
\end{figure*}

\section{Recurrent Event Memory Architecture}
 Let us consider a natural language processing example: 
 \begin{equation}
     Mary\ went\ to\ the\ kitchen. \notag
 \end{equation}
 In this statement, each word is related to the others. First, each word should be stored separately as an event. Memory for this sentence should be triggered by any word. If one hears $Mary$, we will remember that she went to the kitchen. If we hear the word $went$, we know that the subject is Mary instead of someone else. And if we hear $kitchen$, we know that Mary is there and how she got there. Therefore, in the proposed event memory architecture, a recurrent hidden state should connect the words with each other. And this is also the reason why the attention mechanism is so useful in natural language processing. In the encoding phase, each word should be encoded by the current input as well as the previous hidden state. The hidden state update needs first the previous state information. Besides, the read content from the memory is needed to provide long-term information and the current input should provide instantaneous information.

As shown in Figure \ref{fig:Architecture}, at time $t$, given the new sample $x_t$ and previous memory state $h_{t-1}$, three vectors are generated: a query vector $q_t$, a key vector $k_t$ and a value vector $v_t$. The key vector contains the information for future addressing and the value vector contains the information for the content. For example, for the question $Where\  did\ Mary\ go?$, the information in the word $Mary$ and $go$ are used for addressing and the word $kitchen$ is used for the answer. Both vectors form the event at time $t$. At each time, this event (key-value pair) is pushed into the memory. So, the external neural memory is a set of key-value pairs,
\begin{equation}
\begin{aligned}
    \{(k_{t-1}, v_{t-1}&), (k_{t-2}, v_{t-2}), ..., (k_{t-n}, v_{t-n})\ \\
    &|\ k_i\in R^{n_k}, v_i \in R^{n_v}\}
\end{aligned}
\end{equation}
where $n_k$ is the size of the key vector $k_i$ and $n_v$ is the size of the value vector $v_i$, and $n$ is the size of the memory.

The query vector for the read operation is used for addressing the previous event location in the memory. And it is generated by,
\begin{equation}
    q_t = f_q(x_t, h_{t-1}, W^q)
\end{equation}
where $W^q$ is the weights for $f_q(\cdot)$. Then, by comparing the similarity between $q_t$ and keys in the memory, the read vector is defined by the weighted linear combination of the value vectors.
\begin{equation}
\label{read}
    r_t = \sum_{i=1}^n v_{t-i} K(q_t, k_{t-i})
\end{equation}
where $K(\cdot, \cdot)$ is the similarity measure, which here is implemented by Gaussian. We have shown in \cite{Liu2006CorrentropyAL} that the Gaussian creates an induced metric that contains the information of all the even  moments of the probability density of the input data. 

For each event in the memory, the key vector $k_t$ and value vector $v_t$ are generated taking the same input but different parameters.
\begin{equation}
    k_t = f_k(x_t, h_{t-1}, W^k)
\end{equation}
\begin{equation}
    v_t = f_v(x_t, h_{t-1}, W^v)
\end{equation}
where $W^k$ and $W^v$ are the weights for $f_k(\cdot)$ and $f_v(\cdot)$.

Then, the key-value pair is pushed into the memory. When the memory is full, it follows the first-in-first-out (FIFO) rule to discard and add events. The hidden state is updated by,
\begin{equation}
    h_t = f_h(r_t, x_t, h_{t-1}, W^h)
\end{equation}
where $W_h$ is the weights for $f_h(\cdot)$.

The output is determined both by the read content and the hidden state as,
\begin{equation}
    o_t = f_o(r_t, h_{t-1}, W^o)
\end{equation}
where $W_o$ is the weights for $f_o(\cdot)$.
\\
\section{Linear Controller Implementation (MemNet) and Backpropagation Through Time}
The simplest structure for MemNet employs only linear operators for all blocks which is well-understood and much easier to train. To simplify, the sizes for all the functional vectors ($q_t$, $k_t$, $v_t$, $h_t$) are the same and referred as $n_h$. Then the update equations are,
\begin{equation}
    q_t = W_x^q x_t + W^q_h h_{t-1}
\end{equation}
\begin{equation}
    k_t = W_x^k x_t + W^k_h h_{t-1}
\end{equation}
\begin{equation}
    v_t = W_x^v x_t + W^v_h h_{t-1}
\end{equation}
Denote the size for $x_t$ is $n_x$. $W_x^q$, $W_x^k$ and $W_x^v$ are matrices with size $n_h\times n_x$, and $W_h^q$, $W_h^k$ and $W_h^v$ are matrices with size $n_h\times n_h$. In order to implement the memory in coding, we create two memory buffers, $Key_t \in R^{n \times n_h} $ for the key and $Value_t \in R^{n \times n_h} $ for the value, following the first-in-first-out (FIFO) principle. At each time, the new key-value pair is pushed into the memory and the pair at the bottom is discarded. When the memory is empty, the key and value matrices are initialized with zeros. Using the Gaussian similarity, the read content $r_t$ can be expressed as,
\begin{equation}
    r_t = \sum_{i=1}^{n}{v_{t-i} G_\sigma({k_{t-i}, q_t}) }
\label{eq:read}
\end{equation}
where $\sigma$ is the kernel size. And the update for hidden state and output is,
\begin{equation}
    h_t = W_r^h r_t + W_x^h x_t + W^h_h h_{t-1}
\end{equation}
\begin{equation}
    o_t = W_r^o r_t + W^o_h h_{t-1}
\end{equation}
Denote the size for $o_t$ is $n_o$. The sizes for $W_r^h$ and $W^h_h$ are $n_h\times n_h$. The sizes for $W_x^h$ is $n_h\times n_x$ and for $W_r^o$ and $W_h^o$ are $n_o\times n_h$.
From the equations we can see that MemNet can be regarded as an vanilla RNN equipped with the event memory architecture without explicit nonlinear activation function.
Then, the error at time $t$,
\begin{equation}
    e_t = d_t - o_t
\end{equation}

Taking the mean squared error (MSE) as the cost function, at time $t+1$,
\begin{equation}
    J_{t+1}(W) = \frac{1}{2}e_{t+1}^T e_{t+1}
\end{equation}
where $W=\{W_x^q; W_h^q; W_x^k; W_h^k; W_x^v; W_h^v; W_x^h; W_h^h; W_r^h;\\ W_r^o; W_h^o\}$.

When taking the gradient,
\begin{equation}
    \frac{dJ_{t+1}}{dW} = - e_{t+1}^T \frac{do_{t+1}}{dW}
\end{equation}

For the output layer,
\begin{equation}
    \frac{dJ_{t+1}}{dW_r^o} = -  e_{t+1}^T r_{t+1}
\end{equation}
\begin{equation}
    \frac{dJ_{t+1}}{dW_h^o} = -  e_{t+1}^T h_{t}
\end{equation}

For the other blocks, apply the chain rule,

\begin{equation}
     \frac{dJ_{t+1}}{do_{t+1}}\frac{do_{t+1}}{dW}= \frac{dJ_{t+1}}{do_{t+1}} (\frac{do_{t+1}}{dr_{t+1}}\frac{dr_{t+1}}{dW} + \frac{do_{t+1}}{dh_{t}}\frac{dh_{t}}{dW})
\end{equation}
\begin{equation}
    \frac{do_{t+1}}{dr_{t+1}} = W_r^o
\end{equation}
\begin{equation}
    \frac{do_{t+1}}{dh_{t}} = W_h^o
\end{equation}

The gradient along the path of the hidden state $h_{t}$ is computed through time,
\begin{equation}
    \frac{dh_{t}}{dW} = W_r^h \frac{dr_{t}}{dW} + W_h^h \frac{dh_{t-1}}{dW}
\end{equation}

And the gradient to $r_t$ are propagated to the $q_t$, $k_t$ and $v_t$ vectors.
\begin{align}
    &\frac{dr_t}{dq_t} = -\frac{1}{\sigma} \sum_{i=1}^n v_{t-i}(q_t-k_{t-i})e^{-\frac{||q_t - k_{t-i} ||^2}{2\sigma} }\\
    &\frac{dr_t}{dk_{t-i}} = \frac{1}{\sigma} v_{t-i}(q_t-k_{t-i})e^{-\frac{||q_t - k_{t-i} ||^2}{2\sigma} }\\
    &\frac{dr_t}{dv_{t-i}} = e^{-\frac{||q_t - k_{t-i} ||^2}{2\sigma}}
\end{align}

Given the gradients above, the query, key and value vectors are optimized and they tend to learn different types of information for different usages. For instance, the weights for the query vector are more likely to learn long term dependency compared with other two vectors because its gradient is the summation of the list of event memories.
Then, for the read content at time $t$, 
\begin{equation}
    \frac{dr_{t}}{dW} = \frac{dr_t}{dq_t}\frac{dq_t}{dW} + \sum_{i=1}^n(\frac{dr_t}{dk_{t-i}}\frac{dk_{t-i}}{dW} + \frac{dr_t}{dv_{t-i}}\frac{dv_{t-i}}{dW})
\end{equation}

For the read block,
\begin{equation}
    \frac{dq_t}{dW_x^q} = x_t
\end{equation}
\begin{equation}
    \frac{dq_t}{dW_h^q} = h_{t-1}
\end{equation}

For the weights other than $W_x^q$ and $W_h^q$,
\begin{equation}
    \frac{dq_t}{dW'} = W_h^q \frac{dh_{t-1}}{dW'}
\end{equation}

The gradients are the same for key block and value block,
\begin{equation}
    \frac{dk_t}{dW_x^k} = x_t
\end{equation}
\begin{equation}
    \frac{dk_t}{dW_h^k} = h_{t-1}
\end{equation}
\begin{equation}
    \frac{dv_t}{dW_x^v} = x_t
\end{equation}
\begin{equation}
    \frac{dv_t}{dW_h^v} = h_{t-1}
\end{equation}

For the weights from other blocks,
\begin{equation}
    \frac{dk_t}{dW'} = W_h^k \frac{dh_{t-1}}{dW'}
\end{equation}
\begin{equation}
    \frac{dv_t}{dW'} = W_h^v \frac{dh_{t-1}}{dW'}
\end{equation}
\\
\section{Experiment and Results}
\label{sec:Experiment}
In MemNet, we use key-value pairs and external memory to enhance long-term dependency. Since we use Gaussian as the similarity measure, which is localized, the information retrieval should be sparse and it can discard useless information even for recent samples. Therefore, in the first experiment (Chaotic Time Series), we visualize the similarities between the query and keys at each time to see how the information from past samples is preserved and/or discarded. Also, remembering Equation \ref{eq:read}, the read content is the product of a linear projection (value) and a nonlinear similarity measure. This arrangement solves the problems of saturation and vanishing gradient caused by nonlinear activation functions because now the gradient is free from the data scale. We show this benefit in a real word dataset (Air Passenger Time Series). Finally, in the external event memory, the keys are used for addressing and behave like physical locations (indexes). We run the third experiment (Symbolic Operation Tasks), where input sequences are randomly generated and without any temporal information, to visualize how the MemNet index each sample just by the time index. As a comparison, we test MemNet on the question-answering dataset bAbI to show its capability on language understanding and information retrieval.
\subsection{Chaotic Time Series Prediction}
To begin with, we test our new MemNet model on the Hénon map, which is a discrete-time dynamical system that exhibits chaotic behavior. At each time, the system takes the previous coordinate $(x_{t-1}, y_{t-1})$ and maps it to a new coordinate,
\begin{equation}
    \left\{
\begin{aligned}
&x_t = 1-1.4x_{t-1}^2 + y_{t-1} \\
&y_t = 0.3 x_{t-1}
\end{aligned}
\right.
\end{equation}

\begin{figure}[h]
    \centering
    \includegraphics[scale=0.32]{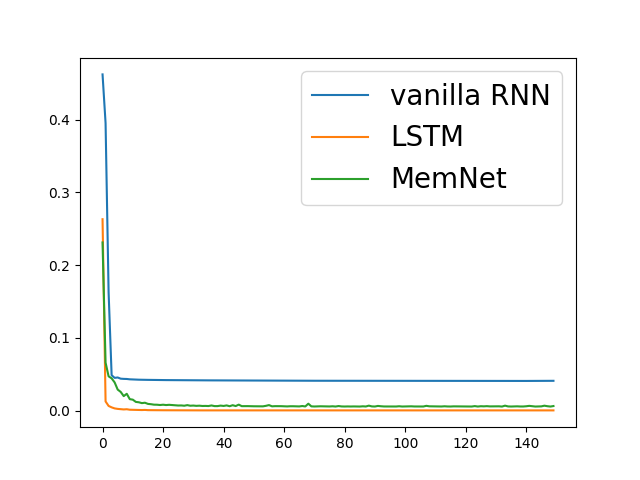}
    \caption{Learning curve on Hénon map.The performance of MemNet is close to LSTM.}
    \label{fig:chaotic}
\end{figure}

In the experiment, we compare the learning curve from a vanilla RNN and LSTM. The hidden layer sizes for the three models are selected by the one that exhibits the best performance. For MemNet, the hidden embedding size is 8, and the external memory size is 64. Here the hidden embedding corresponds to the number of "bits" in the memory, although here each value is a real number. The three models use the same optimizer, Adam with a learning rate of 0.01. The learning curve is shown in Figure \ref{fig:chaotic}. The best performer is the LSTM because of its longer memory, but notice that the MemNet is very close to its performance and it is much simpler to train. They both outperform the vanilla RNN. So we conclude that MemNet is very effective for time series prediction, even though we are just adapting linear mappers. 
\begin{figure}[h]
    \centering
    \includegraphics[scale=0.32]{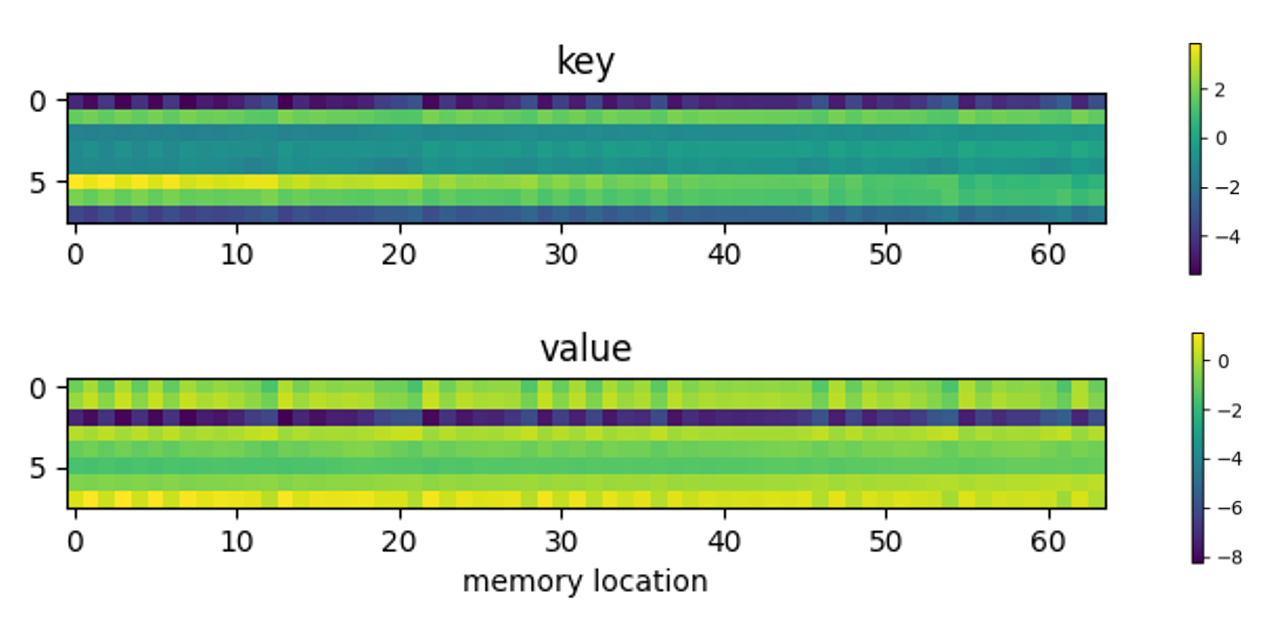}
    \caption{Key and value visualization at the end of the sequence.}
    \label{fig:chaotic key value}
\end{figure}

\begin{figure}[h]
    \centering
    \includegraphics[scale=0.35]{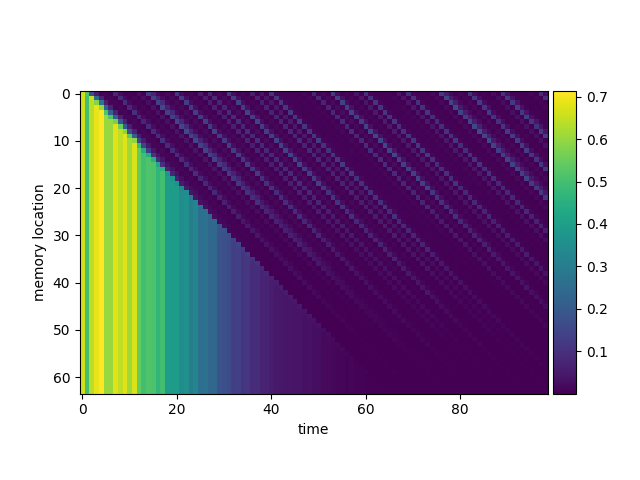}
    \caption{Memory access through time.}
    \label{fig:chaotic read}
\end{figure}

Next, we analyze the internal representations of MemNet. Figure \ref{fig:chaotic key value} plots the key and value matrices in the external event memory at the end of the training sequence. Note that more recent events are at the top of the buffer, near zero at the left of the X-axis. The Y axis represents the 8 embedding dimensions. As expected, the values of the memory keys are evolving smoothly across training to provide the required memory depth for addressing. However, the "bits" for the value sub-module quickly change  from sample to sample across the training to provide accurate information resolution for each memory location.
% It shows that the keys fade away from left to right, which is the inverse order of time. No keys at two different times are the same and they are only used for addressing. Note that the value has a drastically different content.
 Figure \ref{fig:chaotic read} provides a visualization of the memory access (similarity measure between the query vector and the keys) through testing.  There are 64 locations in the memory so the similarity measure should be a vector with size 64 at each time. A high value means the corresponding weight for the value is high and a strong connection between the past sample and the current input is made. New memory values are pushed one by one and the one at point 0 on Y axis is the latest memory item. Since the memory is initialized with zeros for the key and value matrices, the similarities are constant for these locations when the memory is not full (vertical bars in the left part of the memory buffer). Also, due to the lack of information from hidden states and the read content, the queries at the beginning are close to zero, causing high similarities with zero keys. And the query vector is away from zero when building the memory and the similarities become zero. Theoretically, the similarity is more discriminant in the quasi-linear region of the Gaussian where Gaussian has the largest slope. In practice, the keys and queries tend to fall into exactly this region as we can verify by the color code. So we can expect that in regression applications, the similarity can't be close to 1. The same Figure \ref{fig:chaotic read} shows that once the memory is full, the information is retrieved quasi-periodically through time, mimicking the time structure of the Hennon time series. Note that the diagonal lines appear because the memory in the buffer is always shifted one right i.e., the same information appears in the next column.

\subsection{Real World Non-Stationary Time Series}

To validate the assumption that MemNet is free from the scale of the data, we test the performance on the Air Passenger Dataset, which contains the number of monthly air passengers in thousands from 1949 until 1960. The difficulty in predicting this data set is that the mean and variance tend to increase over time, so it is a heteroscedastic time series. The model has to learn the mean trend as well as the periodic seasonal fluctuation. 
\begin{figure}[h]
    \centering
    \includegraphics[scale=0.42]{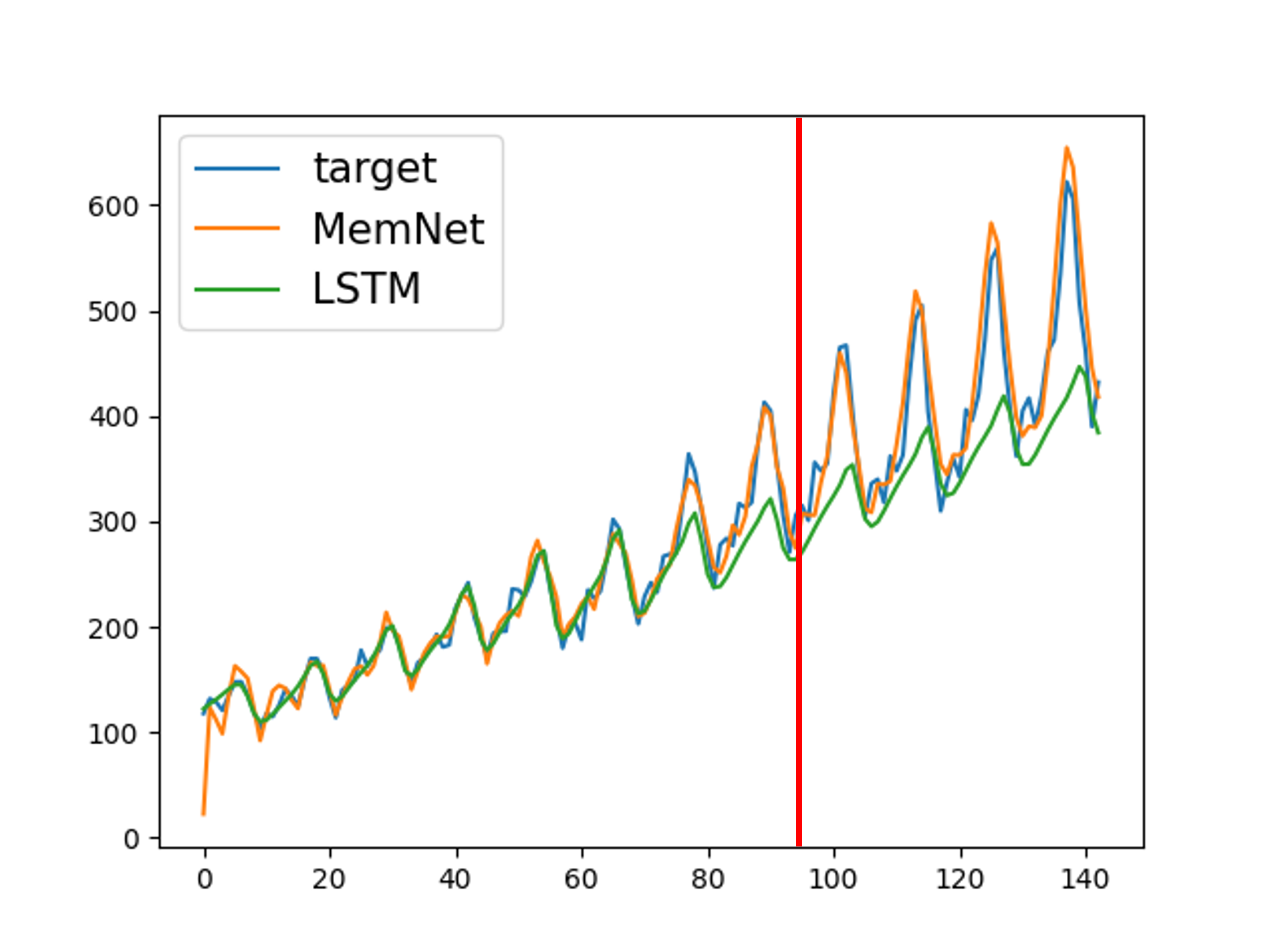}
    \caption{Task 1: Recursive prediction result. The left of the red line are the training set and the right of the red line are the testing set.}
    \label{fig:long}
\end{figure}
\begin{figure}[h]
    \centering
    \includegraphics[scale=0.32]{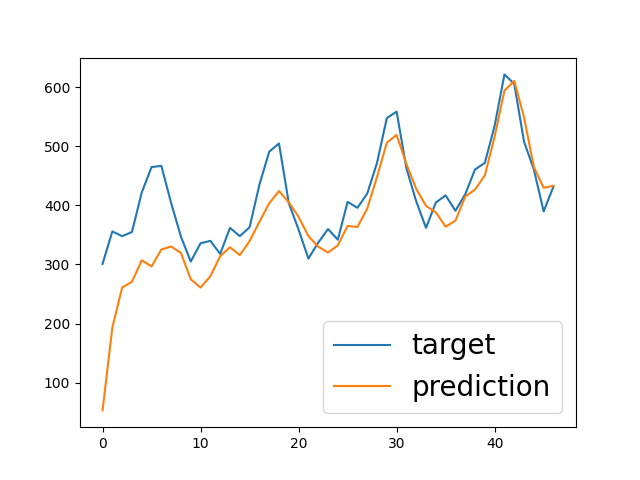}
    \caption{Task 2: Synchronize task. The model synchronizes after the event memory has been built.}
    \label{fig:syn}
\end{figure}

In the experiment, there are 144 data points in the dataset and we use the first 96 points as the training set. We also take the last 24 points in the training set as the validation set. The model order is selected by the one that has the smallest error on the validation data. The embedding size is 12 and the external memory size is 16. We also choose Adam as our optimizer and the learning rate is 0.01. When the model is well-trained, we run two different types of tasks on the test set. The first task is to test the long-term prediction performance. In this task, the training data is fed into the model to create event memories. Then, it uses the previous prediction as the next input, not the input from the data set. The model is going to predict the later 48 test points only using the information contained in the external memory. And the result is shown in Figure \ref{fig:long}. We also test the performance of LSTM on the same dataset as a comparison. Because of the heteroscedastic in the time series the conventional LSTM saturates the external memory and can not be used. We take out the mean in the training set data as a pre-processing step and let the LSTM learn just the variance. For testing we add the mean back to the prediction. As shown in Figure \ref{fig:long}, our proposed network learns very well both the trend and fluctuation while the LSTM is incapable of predicting recursively the time series.  

The second task is to test if the MemNet with all parameters fixed from the training set can capture the phase and synchronize with the test data when the external memory is cleared. In this task, the model only takes the test data, fills incrementally the memory and predicts the next value. Figure \ref{fig:syn} shows the result for the second task. MemNet displays poor performance on the prediction during the filling up of the memory, but after 16 samples it start to predict the time series progressively well.

\subsection{Symbolic Operation Tasks}
\begin{figure}[h]
    \centering
    \includegraphics[scale=0.44]{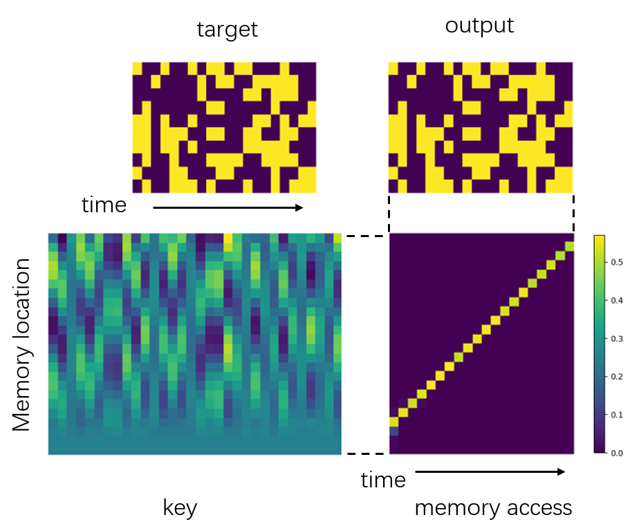}
    \caption{Copy task visualization.}
    \label{fig:copy task}
\end{figure}

\begin{figure}[h]
    \centering
    \includegraphics[scale=0.44]{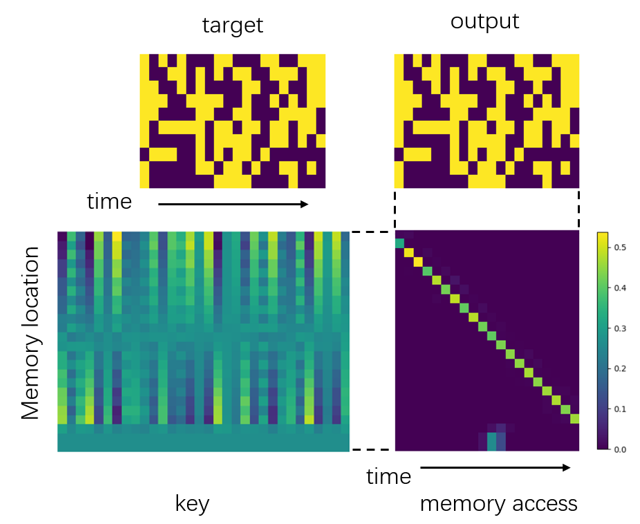}
    \caption{Reverse task visualization.}
    \label{fig:reverse task}
\end{figure}

To visualize how the MemNet works on symbolic operation tasks in bit strings such as copy and reverse character, we compare its performance with the Neural Turing Machine \cite{graves2014neural}. The input data is a sequence of bits generated randomly and the goal is to return the same symbols in the same order or its reverse order. For each task, we train and test MemNet on sequences with lengths from 1 to 20. The embedding size is 32 and event memory size is 128. The optimizer is Adam with a learning rate of 0.0001. The visualizations for the two tasks are shown in Figures \ref{fig:copy task} and \ref{fig:reverse task}. The testing results show that the MemNet performs the tasks with no errors for strings up to 20. It is interesting to see that the learned memory keys in the external neural memory for the same input are quite different in the two tasks. This can be expected because the targets are quite different so the operations must also be different. However, it is interesting that the memory access is the mirror image between the two tasks, which can be expected since the order of the samples is flipped between the tasks. Note that for this task, MemNet relies more on the keys to capture order information and the similarity is higher than the Henon Map experiment. The final experiment is to test the generalization of MemNet in the copy task for longer sequences.  

\begin{figure}[h]
    \centering
    \includegraphics[scale=0.45]{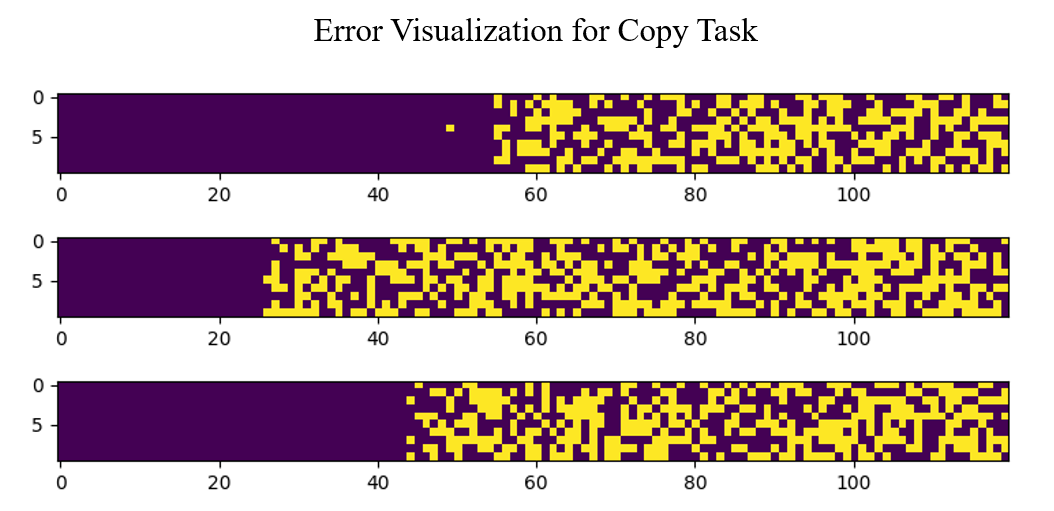}
    \caption{Copy task error for longer sequences. Visualization for three random sequences. Purple means the prediction is correct and yellows means the prediction is wrong at that location.}
    \label{fig:copy}
\end{figure}
In order to test the generalization of the model for longer sequences, we use the same model trained above. We randomly generate sequences with the length of 120 and plot three of them. Figure \ref{fig:copy} shows that our model can generalize for longer sequences but starts failing after length 25 for some of the sequences, while for others it can generalize for lengths up to 50. We could not relate the structure of the sequence with the generalization.

\subsection{Question Answering Dataset (bAbI)}
In the last experiment, we test the performance of MemNet on the synthetic question-answering dataset, bAbI \cite{weston2015towards}. We have to say that the difference in complexity of MemNet versus the currently employed models in bAbI makes the comparison a bit unfair, but we decided to proceed just to show the potential of MemNet to solve problems that currently require much more involved computation and training times. The dataset contains 20 different types of questions, which are designed for text understanding and reasoning. Each task contains a list of stories. For each story, there are several clue sentences mixed with some irrelevant sentences. And each story has several questions. For example,
\begin{align}
&1\ Mary\ got\ the\ football\ there. \notag\\ 
&2\ John\ went\ to\ the\ kitchen.\notag \\ 
&3\ Mary\ went\ back\ to\ the\ kitchen.\notag \\
&4\ Mary\ went\ back\ to\ the\ garden.\notag\\
&5\ Where\ is\ the\ football?\ 	garden\	1\ 4\notag\\
&6\ Daniel\ went\ back\ to\ the\ kitchen.\notag \\
&7\ Mary\ dropped\ the\ football.\notag \\
&8\ John\ got\ the\ milk\ there.\notag\\
&9\ Where\ is\ the\ football?\ 	garden\	7\ 4\notag
\end{align}

The first number in each sentence is the index. And the number after the question is the reference for the question. When answering the question for sentence 5, $Where\ is\ the\ football?$, the useful information contains in sentence 1 and sentence 4. This task is called 2 supporting facts because for each question there are two references.

In the experiment, each story is a separate sequence. And every word and punctuation is a vector using the one-hot encoding. No other word embedding techniques are used. For each sequence, we follow the order of the statements and questions. MemNet's memory is reset at the beginning of each sequence. When it encounters a question, the model stops memorizing and takes the question as input. The last output of the model is treated as the answer. After answering this question, the model continues memorizing the new statements for future questions. For the experiment, we run two types of training on the dataset, single and joint queries. For single training, 20 models are trained separately on the 20 tasks. The memory size for the model is 512. And the hidden size is uniformly set to 32. We use a single linear layer for all the blocks in Figure \ref{fig:Architecture}. Since the model stores the encoded information and the goal is to extract the correct information for the answer, this task is not a classification problem. Therefore, no softmax output is used and we choose the MSE as the objective function for simplicity. We use Adam as the optimizer with a learning rate of 0.0001 and no weight decay. No zero padding is used for online learning. For the joint task, there are 194 words \& punctuation in total and the size for the one-hot embedding is 194. We use the same MemNet configuration but use 64 as the embedding size. These results should be considered preliminary due to the lack of computational resources and restricted training time. We report the best result in 10 training runs. MemNet achieves a mean error rate of 2.96\% and 1 failed task for single training; and a mean error rate of 5.6\% with 3 failures for joint training (Table \ref{tab1}). 
\begin{table*}[h]
\caption{bAbI results comparison. MEMO \cite{banino2020memo} is the current state of art model using transformer.  And all other models (except MemN2N) are recurrent neural networks equipped with external memories. The result of NTM and DNC are from \cite{graves2016hybrid}. The results of MemN2N come from \cite{sukhbaatar2015end}. The result of DMN is from \cite{kumar2016ask}.}
\begin{center}
\begin{adjustbox}{width=1\textwidth}
\begin{tabular}{|l|cccccccc|}
\hline
\multicolumn{1}{|c|}{\multirow{3}{*}{Task}} & \multicolumn{8}{c|}{bAbI Best Results}    \\ \cline{2-9} 
\multicolumn{1}{|c|}{}                      & \multicolumn{1}{c|}{NTM}     & \multicolumn{1}{c|}{DNC}     & \multicolumn{1}{c|}{MemN2N}  & \multicolumn{1}{c|}{MemN2N}   & \multicolumn{1}{c|}{DMN}      & \multicolumn{1}{c|}{MEMO*} & \multicolumn{1}{c|}{MemNet}   & MemNet  \\
\multicolumn{1}{|c|}{}                      & \multicolumn{1}{c|}{(Joint)} & \multicolumn{1}{c|}{(Joint)} & \multicolumn{1}{c|}{(Joint)} & \multicolumn{1}{c|}{(Single)} & \multicolumn{1}{c|}{(Single)} & \multicolumn{1}{c|}{(Joint)}           & \multicolumn{1}{c|}{(Single)} & (Joint) \\ \hline
1: 1   supporting fact                      & \multicolumn{1}{c|}{31.5}    & \multicolumn{1}{c|}{0}       & \multicolumn{1}{c|}{0}       & \multicolumn{1}{c|}{0}        & \multicolumn{1}{c|}{0}        & \multicolumn{1}{c|}{0}                 & \multicolumn{1}{c|}{0}        &
\multicolumn{1}{c|}{0}\\
2: 2 supporting facts                       & \multicolumn{1}{c|}{54.5}    & \multicolumn{1}{c|}{1.3}     & \multicolumn{1}{c|}{1}       & \multicolumn{1}{c|}{0.3}      & \multicolumn{1}{c|}{1.8}      & \multicolumn{1}{c|}{0}                 & \multicolumn{1}{c|}{1.6}      &
\multicolumn{1}{c|}{0.17}\\
3: 3 supporting facts                       & \multicolumn{1}{c|}{43.9}    & \multicolumn{1}{c|}{2.4}     & \multicolumn{1}{c|}{6.8}     & \multicolumn{1}{c|}{2.1}      & \multicolumn{1}{c|}{4.8}      & \multicolumn{1}{c|}{2.95}              & \multicolumn{1}{c|}{2.6}      &
\multicolumn{1}{c|}{0.7}\\
4: 2 argument rels                          & \multicolumn{1}{c|}{0}       & \multicolumn{1}{c|}{0}       & \multicolumn{1}{c|}{0}       & \multicolumn{1}{c|}{0}        & \multicolumn{1}{c|}{0}        & \multicolumn{1}{c|}{0}                 & \multicolumn{1}{c|}{0}      &
\multicolumn{1}{c|}{0}\\
5: 3 argument rels                          & \multicolumn{1}{c|}{0.8}     & \multicolumn{1}{c|}{0.5}     & \multicolumn{1}{c|}{6.1}     & \multicolumn{1}{c|}{0.8}      & \multicolumn{1}{c|}{0.7}      & \multicolumn{1}{c|}{0}                 & \multicolumn{1}{c|}{1.3}      &
\multicolumn{1}{c|}{0.8}\\
6: yes/no questions                         & \multicolumn{1}{c|}{17.1}    & \multicolumn{1}{c|}{0}       & \multicolumn{1}{c|}{0.1}     & \multicolumn{1}{c|}{0.1}      & \multicolumn{1}{c|}{0}        & \multicolumn{1}{c|}{0}                 & \multicolumn{1}{c|}{0}        & 
\multicolumn{1}{c|}{0}\\
7: counting                                 & \multicolumn{1}{c|}{17.8}    & \multicolumn{1}{c|}{0.2}     & \multicolumn{1}{c|}{6.6}     & \multicolumn{1}{c|}{2}        & \multicolumn{1}{c|}{3.1}      & \multicolumn{1}{c|}{0}                 & \multicolumn{1}{c|}{0}      & 
\multicolumn{1}{c|}{3.2}\\
8: lists/sets                               & \multicolumn{1}{c|}{13.8}    & \multicolumn{1}{c|}{0.1}     & \multicolumn{1}{c|}{2.7}     & \multicolumn{1}{c|}{0.9}      & \multicolumn{1}{c|}{3.5}      & \multicolumn{1}{c|}{0}                 & \multicolumn{1}{c|}{0}        &
\multicolumn{1}{c|}{7.2}\\
9: simple negation                          & \multicolumn{1}{c|}{16.4}    & \multicolumn{1}{c|}{0}       & \multicolumn{1}{c|}{0}       & \multicolumn{1}{c|}{0.3}      & \multicolumn{1}{c|}{0}        & \multicolumn{1}{c|}{0}                 & \multicolumn{1}{c|}{0}        &
\multicolumn{1}{c|}{0}\\
10: indefinite knowl.                       & \multicolumn{1}{c|}{16.6}    & \multicolumn{1}{c|}{0.2}     & \multicolumn{1}{c|}{0.5}     & \multicolumn{1}{c|}{0}        & \multicolumn{1}{c|}{0}        & \multicolumn{1}{c|}{0}                 & \multicolumn{1}{c|}{0.1}      &
\multicolumn{1}{c|}{0.6}\\
11: basic coreference                       & \multicolumn{1}{c|}{15.2}    & \multicolumn{1}{c|}{0}       & \multicolumn{1}{c|}{0}       & \multicolumn{1}{c|}{0.1}      & \multicolumn{1}{c|}{0.1}      & \multicolumn{1}{c|}{0}                 & \multicolumn{1}{c|}{0}        &
\multicolumn{1}{c|}{0}\\
12: conjunction                             & \multicolumn{1}{c|}{8.9}     & \multicolumn{1}{c|}{0.1}     & \multicolumn{1}{c|}{0.1}     & \multicolumn{1}{c|}{0}        & \multicolumn{1}{c|}{0}        & \multicolumn{1}{c|}{0}                 & \multicolumn{1}{c|}{0}        &
\multicolumn{1}{c|}{0}\\
13: compound coref.                         & \multicolumn{1}{c|}{7.4}     & \multicolumn{1}{c|}{0}       & \multicolumn{1}{c|}{0}       & \multicolumn{1}{c|}{0}        & \multicolumn{1}{c|}{0.2}      & \multicolumn{1}{c|}{0}                 & \multicolumn{1}{c|}{0}      &
\multicolumn{1}{c|}{0}\\
14: time reasoning                          & \multicolumn{1}{c|}{24.2}    & \multicolumn{1}{c|}{0.3}     & \multicolumn{1}{c|}{0}       & \multicolumn{1}{c|}{0.1}      & \multicolumn{1}{c|}{0}        & \multicolumn{1}{c|}{0}                 & \multicolumn{1}{c|}{0}      &
\multicolumn{1}{c|}{4.2}\\
15: basic deduction                         & \multicolumn{1}{c|}{47}      & \multicolumn{1}{c|}{0}       & \multicolumn{1}{c|}{0.2}     & \multicolumn{1}{c|}{0}        & \multicolumn{1}{c|}{0}        & \multicolumn{1}{c|}{0}                 & \multicolumn{1}{c|}{0}        &
\multicolumn{1}{c|}{0}\\
16: basic induction                         & \multicolumn{1}{c|}{53.6}    & \multicolumn{1}{c|}{52.4}    & \multicolumn{1}{c|}{0.2}     & \multicolumn{1}{c|}{51.8}     & \multicolumn{1}{c|}{0.6}      & \multicolumn{1}{c|}{1.25}              & \multicolumn{1}{c|}{48.5}     &
\multicolumn{1}{c|}{0.6}\\
17: positional reas.                        & \multicolumn{1}{c|}{25.5}    & \multicolumn{1}{c|}{24.1}    & \multicolumn{1}{c|}{41.8}    & \multicolumn{1}{c|}{18.6}     & \multicolumn{1}{c|}{40.4}     & \multicolumn{1}{c|}{0}                 & \multicolumn{1}{c|}{0}        &
\multicolumn{1}{c|}{6.3}\\
18: size reasoning                          & \multicolumn{1}{c|}{2.2}     & \multicolumn{1}{c|}{4}       & \multicolumn{1}{c|}{8}       & \multicolumn{1}{c|}{5.3}      & \multicolumn{1}{c|}{4.7}      & \multicolumn{1}{c|}{0}                 & \multicolumn{1}{c|}{0.1}      &
\multicolumn{1}{c|}{0}\\
19: path finding                            & \multicolumn{1}{c|}{4.3}     & \multicolumn{1}{c|}{0.1}     & \multicolumn{1}{c|}{75.7}    & \multicolumn{1}{c|}{2.3}      & \multicolumn{1}{c|}{65.5}     & \multicolumn{1}{c|}{0}                 & \multicolumn{1}{c|}{4.7}       &
\multicolumn{1}{c|}{87.3}\\
20: agent motiv.                            & \multicolumn{1}{c|}{1.5}     & \multicolumn{1}{c|}{0}       & \multicolumn{1}{c|}{0}       & \multicolumn{1}{c|}{0}        & \multicolumn{1}{c|}{0}        & \multicolumn{1}{c|}{0}                 & \multicolumn{1}{c|}{0}        &        
\multicolumn{1}{c|}{0}        \\ \hline
Mean Err. (\%)                           &  \multicolumn{1}{c|}{20.1}       & \multicolumn{1}{c|}{4.3}       & \multicolumn{1}{c|}{7.5}        & \multicolumn{1}{c|}{4.2}        & \multicolumn{1}{c|}{6.4}                 & \multicolumn{1}{c|}{0.21}        &  
\multicolumn{1}{c|}{2.96}     &
\multicolumn{1}{c|}{5.6}        \\ 
Failed (err.\textgreater5\%)                    & \multicolumn{1}{c|}{16}     & \multicolumn{1}{c|}{2}       & \multicolumn{1}{c|}{6}       & \multicolumn{1}{c|}{3}        & \multicolumn{1}{c|}{2}        & \multicolumn{1}{c|}{0}                 & \multicolumn{1}{c|}{1}        &        
\multicolumn{1}{c|}{3}        \\ \hline
\end{tabular}
\label{tab1}
\end{adjustbox}
\end{center}
\end{table*}
MemNet provides high performance in most of the tasks but performance degrades in the joint task as can be expected. However, it is hard to interpret why sometimes MemNet fails miserably (e.g. task 16 in single mode and in task 19 in the joint mode), while in the alternative mode performance is good. We think that this is associated with insufficient training. MemNet results are worse than the state-of-the-art transformer network, but our architecture and training is much simpler. When we compare the MemNet with the NTM and DNC that also use state models with external neural memory, we observe that MemNet mean error is much better than NTM and slightly worse than the DNC, which are both much more complex architectures. MemNet compares favorably with the attention networks MemN2N and DMN, which are also quite large networks. In fact, as we can see from Table \ref{tab:par}, MemNet with the hidden size of 64 
\begin{table}[t]
\caption{Comparison of hyperparameters with other recurrent models. The hidden size for NTM and DNC are referred as the hidden size for LSTM controller. For the model size, we roughly calculate the number of trainable parameters and assume all LSTMs are single layer in \cite{graves2016hybrid}. }
\begin{adjustbox}{width=0.46\textwidth}
\begin{tabular}{|c|c|c|c|c|}
\hline
 & LSTM    & NTM         & DNC      & MemNet \\ \hline
Hidden Size   & 512    & 256        & 256     & 64     \\
Learning Rate   & $10^{-4}$      &     $10^{-4}$         & $10^{-4}$         & $10^{-4}$  \\
Read Heads         & $\sim$           & 16         & 16       & 1      \\
\begin{tabular}[c]{@{}c@{}}Model Size\end{tabular} & \multicolumn{1}{l|}{\textgreater{}1452k} & \multicolumn{1}{l|}{\textgreater{}846k} & \textgreater{}1050k & 95k    \\ \hline
\end{tabular}
\end{adjustbox}
\label{tab:par}
\end{table}

has a much smaller number of trainable parameters than the others on the jointly training task, therefore it is surprising that it can compete with other complex architectures. Since we use linear models in MemNet, training is very stable and converges faster.

\section{Conclusion}
In this paper, we propose a universal recurrent event memory architecture  (MemNet) for streaming data. The key innovation is to separate the event information for addressing and for content and store them in an external memory, which gets rid of the trade-off between the resolution and memory depth. The key-value pairs greatly simplify the construction of the nonlinearity that is necessary for memory application leading to the use of linear adaptive mapping functions. Thus, the memory should also be decoded by one linear layer, instead of using softmax output, which is commonly used in other state of art models in NLP. In the experiment, we test MemNet on normal time series as well as natural language question-answering tasks. It is gratifying to see that the same architecture can be applied to such different types of applications, unlike the current memory models. Our preliminary results show that in spite of the fact that MemNet has a complexity much smaller than natural language processing networks it can compete with them in terms of accuracy. Therefore, we think that this model will have a niche in IoT for both engineering and natural language processing applications.

\bibliographystyle{IEEEtranS}
\bibliography{ref}

% Generated by IEEEtranS.bst, version: 1.12 (2007/01/11)
\begin{thebibliography}{10}
\providecommand{\url}[1]{#1}
\csname url@samestyle\endcsname
\providecommand{\newblock}{\relax}
\providecommand{\bibinfo}[2]{#2}
\providecommand{\BIBentrySTDinterwordspacing}{\spaceskip=0pt\relax}
\providecommand{\BIBentryALTinterwordstretchfactor}{4}
\providecommand{\BIBentryALTinterwordspacing}{\spaceskip=\fontdimen2\font plus
\BIBentryALTinterwordstretchfactor\fontdimen3\font minus
  \fontdimen4\font\relax}
\providecommand{\BIBforeignlanguage}[2]{{%
\expandafter\ifx\csname l@#1\endcsname\relax
\typeout{** WARNING: IEEEtranS.bst: No hyphenation pattern has been}%
\typeout{** loaded for the language `#1'. Using the pattern for}%
\typeout{** the default language instead.}%
\else
\language=\csname l@#1\endcsname
\fi
#2}}
\providecommand{\BIBdecl}{\relax}
\BIBdecl

\bibitem{akaike1973maximum}
H.~Akaike, ``Maximum likelihood identification of gaussian autoregressive
  moving average models,'' \emph{Biometrika}, vol.~60, no.~2, pp. 255--265,
  1973.

\bibitem{banino2020memo}
A.~Banino, A.~P. Badia, R.~K{\"o}ster, M.~J. Chadwick, V.~Zambaldi,
  D.~Hassabis, C.~Barry, M.~Botvinick, D.~Kumaran, and C.~Blundell, ``Memo: A
  deep network for flexible combination of episodic memories,'' \emph{arXiv
  preprint arXiv:2001.10913}, 2020.

\bibitem{benjamin2003generalized}
M.~A. Benjamin, R.~A. Rigby, and D.~M. Stasinopoulos, ``Generalized
  autoregressive moving average models,'' \emph{Journal of the American
  Statistical association}, vol.~98, no. 461, pp. 214--223, 2003.

\bibitem{cho2014properties}
K.~Cho, B.~Van~Merri{\"e}nboer, D.~Bahdanau, and Y.~Bengio, ``On the properties
  of neural machine translation: Encoder-decoder approaches,'' \emph{arXiv
  preprint arXiv:1409.1259}, 2014.

\bibitem{de1992gamma}
B.~De~Vries and J.~C. Principe, ``The gamma model—a new neural model for
  temporal processing,'' \emph{Neural networks}, vol.~5, no.~4, pp. 565--576,
  1992.

\bibitem{giles1992learning}
C.~L. Giles, C.~B. Miller, D.~Chen, H.-H. Chen, G.-Z. Sun, and Y.-C. Lee,
  ``Learning and extracting finite state automata with second-order recurrent
  neural networks,'' \emph{Neural Computation}, vol.~4, no.~3, pp. 393--405,
  1992.

\bibitem{graves2014neural}
A.~Graves, G.~Wayne, and I.~Danihelka, ``Neural turing machines,'' \emph{arXiv
  preprint arXiv:1410.5401}, 2014.

\bibitem{graves2016hybrid}
A.~Graves, G.~Wayne, M.~Reynolds, T.~Harley, I.~Danihelka,
  A.~Grabska-Barwi{\'n}ska, S.~G. Colmenarejo, E.~Grefenstette, T.~Ramalho,
  J.~Agapiou \emph{et~al.}, ``Hybrid computing using a neural network with
  dynamic external memory,'' \emph{Nature}, vol. 538, no. 7626, pp. 471--476,
  2016.

\bibitem{grefenstette2015learning}
E.~Grefenstette, K.~M. Hermann, M.~Suleyman, and P.~Blunsom, ``Learning to
  transduce with unbounded memory,'' \emph{Advances in neural information
  processing systems}, vol.~28, 2015.

\bibitem{hochreiter1997long}
S.~Hochreiter and J.~Schmidhuber, ``Long short-term memory,'' \emph{Neural
  computation}, vol.~9, no.~8, pp. 1735--1780, 1997.

\bibitem{doi:10.1073/pnas.79.8.2554}
\BIBentryALTinterwordspacing
J.~J. Hopfield, ``Neural networks and physical systems with emergent collective
  computational abilities.'' \emph{Proceedings of the National Academy of
  Sciences}, vol.~79, no.~8, pp. 2554--2558, 1982. [Online]. Available:
  \url{https://www.pnas.org/doi/abs/10.1073/pnas.79.8.2554}
\BIBentrySTDinterwordspacing

\bibitem{kumar2016ask}
A.~Kumar, O.~Irsoy, P.~Ondruska, M.~Iyyer, J.~Bradbury, I.~Gulrajani, V.~Zhong,
  R.~Paulus, and R.~Socher, ``Ask me anything: Dynamic memory networks for
  natural language processing,'' in \emph{International conference on machine
  learning}.\hskip 1em plus 0.5em minus 0.4em\relax PMLR, 2016, pp. 1378--1387.

\bibitem{Liu2006CorrentropyAL}
W.~Liu, P.~P. Pokharel, and J.~C. Pr{\'i}ncipe, ``Correntropy: A localized
  similarity measure,'' \emph{The 2006 IEEE International Joint Conference on
  Neural Network Proceedings}, pp. 4919--4924, 2006.

\bibitem{ma2019taxonomy}
Y.~Ma and J.~C. Principe, ``A taxonomy for neural memory networks,'' \emph{IEEE
  transactions on neural networks and learning systems}, vol.~31, no.~6, pp.
  1780--1793, 2019.

\bibitem{193206}
J.~Principe, B.~de~Vries, and P.~de~Oliveira, ``The gamma-filter-a new class of
  adaptive iir filters with restricted feedback,'' \emph{IEEE Transactions on
  Signal Processing}, vol.~41, no.~2, pp. 649--656, 1993.

\bibitem{hmm}
L.~Rabiner and B.~Juang, ``An introduction to hidden markov models,''
  \emph{IEEE ASSP Magazine}, vol.~3, no.~1, pp. 4--16, 1986.

\bibitem{schuller2003hidden}
B.~Schuller, G.~Rigoll, and M.~Lang, ``Hidden markov model-based speech emotion
  recognition,'' in \emph{2003 IEEE International Conference on Acoustics,
  Speech, and Signal Processing, 2003. Proceedings.(ICASSP'03).}, vol.~2.\hskip
  1em plus 0.5em minus 0.4em\relax Ieee, 2003, pp. II--1.

\bibitem{sukhbaatar2015end}
S.~Sukhbaatar, J.~Weston, R.~Fergus \emph{et~al.}, ``End-to-end memory
  networks,'' \emph{Advances in neural information processing systems},
  vol.~28, 2015.

\bibitem{sun2017neural}
G.-Z. Sun, C.~L. Giles, H.-H. Chen, and Y.-C. Lee, ``The neural network
  pushdown automaton: Model, stack and learning simulations,'' \emph{arXiv
  preprint arXiv:1711.05738}, 2017.

\bibitem{vaswani2017attention}
A.~Vaswani, N.~Shazeer, N.~Parmar, J.~Uszkoreit, L.~Jones, A.~N. Gomez,
  {\L}.~Kaiser, and I.~Polosukhin, ``Attention is all you need,''
  \emph{Advances in neural information processing systems}, vol.~30, 2017.

\bibitem{weston2015towards}
J.~Weston, A.~Bordes, S.~Chopra, A.~M. Rush, B.~Van~Merri{\"e}nboer, A.~Joulin,
  and T.~Mikolov, ``Towards ai-complete question answering: A set of
  prerequisite toy tasks,'' \emph{arXiv preprint arXiv:1502.05698}, 2015.

\end{thebibliography}

\end{document}